\documentclass[sigconf]{arxiv}
\input{header.sty}

\renewcommand\footnotetextcopyrightpermission[1]{}
\usepackage{cleveref}
\crefname{section}{§}{§§}
\Crefname{section}{§}{§§}
\begin{document}

\title{Arbitrary Bit-width Network: A Joint Layer-Wise Quantization and Adaptive Inference Approach}

\author{Chen Tang}
\email{tc20@mails.tsinghua.edu.cn}
\affiliation{%
  \institution{Tsinghua University}
  \country{China}
}

\author{Haoyu Zhai}
\email{dhy21@mails.tsinghua.edu.cn}
\affiliation{%
  \institution{Tsinghua University}
  \country{China}
}

\author{Kai Ouyang}
\email{oyk20@mails.tsinghua.edu.cn}
\affiliation{%
  \institution{Tsinghua University}
  \country{China}
}

\author{Zhi Wang}
\email{wangzhi@sz.tsinghua.edu.cn}
\affiliation{%
  \institution{Tsinghua University}
  \country{China}
}

\author{Yifei Zhu}
\email{yifei.zhu@sjtu.edu.cn}
\affiliation{%
  \institution{Shanghai Jiao Tong University}
  \country{China}
}

\author{Wenwu Zhu}
\email{wwzhu@tsinghua.edu.cn}
\affiliation{%
  \institution{Tsinghua University}
  \country{China}
}


\DeclarePairedDelimiter{\nint}\lfloor\rceil
\DeclarePairedDelimiter{\abs}\lvert\rvert

\begin{abstract}
Conventional model quantization methods use a fixed quantization scheme to different data samples, which ignores the inherent ``recognition difficulty'' differences between various samples.
We propose to feed different data samples with varying quantization schemes to achieve a data-dependent dynamic inference, at a fine-grained layer level. 
However, enabling this adaptive inference with changeable layer-wise quantization schemes is challenging because the combination of bit-widths and layers is growing exponentially, making it extremely difficult to train a single model in such a vast searching space and use it in practice. 
To solve this problem, we present the \textbf{A}rbitrary \textbf{B}it-width \textbf{N}etwork (ABN), where the bit-widths of a single deep network can change at runtime for different data samples, with a layer-wise granularity. 
Specifically, first we build a weight-shared layer-wise quantizable ``super-network'' in which each layer can be allocated with multiple bit-widths and thus quantized differently on demand. 
The super-network provides a considerably large number of combinations of bit-widths and layers, each of which can be used during inference \textit{without retraining or storing myriad models}.
Second, based on the well-trained super-network, each layer's runtime bit-width selection decision is modeled as a Markov Decision Process (MDP) and solved by an adaptive inference strategy accordingly. 
Experiments show that the super-network can be built without accuracy degradation, and the bit-widths allocation of each layer can be adjusted to deal with various inputs on the fly. 
On ImageNet classification, we achieve 1.1\% top1 accuracy improvement while saving 36.2\% BitOps. 
\end{abstract}
\maketitle
\section{Introduction}
Model quantization is one of the most promising compression methods for deploying deep neural networks on resource-limited devices. 
It leverages the intrinsic robustness of neural networks in preserving their expressiveness even after reducing their bit-width. 
For example, an 8-bit network can typically increase the inference speed by 4$\times$ compared to a full-precision (32-bit) network, with 4$\times$ less storage space and negligible accuracy degradation \cite{intel2020}. 
Classical quantization methods can be divided into two categories, independent methods and joint methods. 
\begin{figure}[t]
\centering
\includegraphics[height=7.0cm,width=8.0cm]{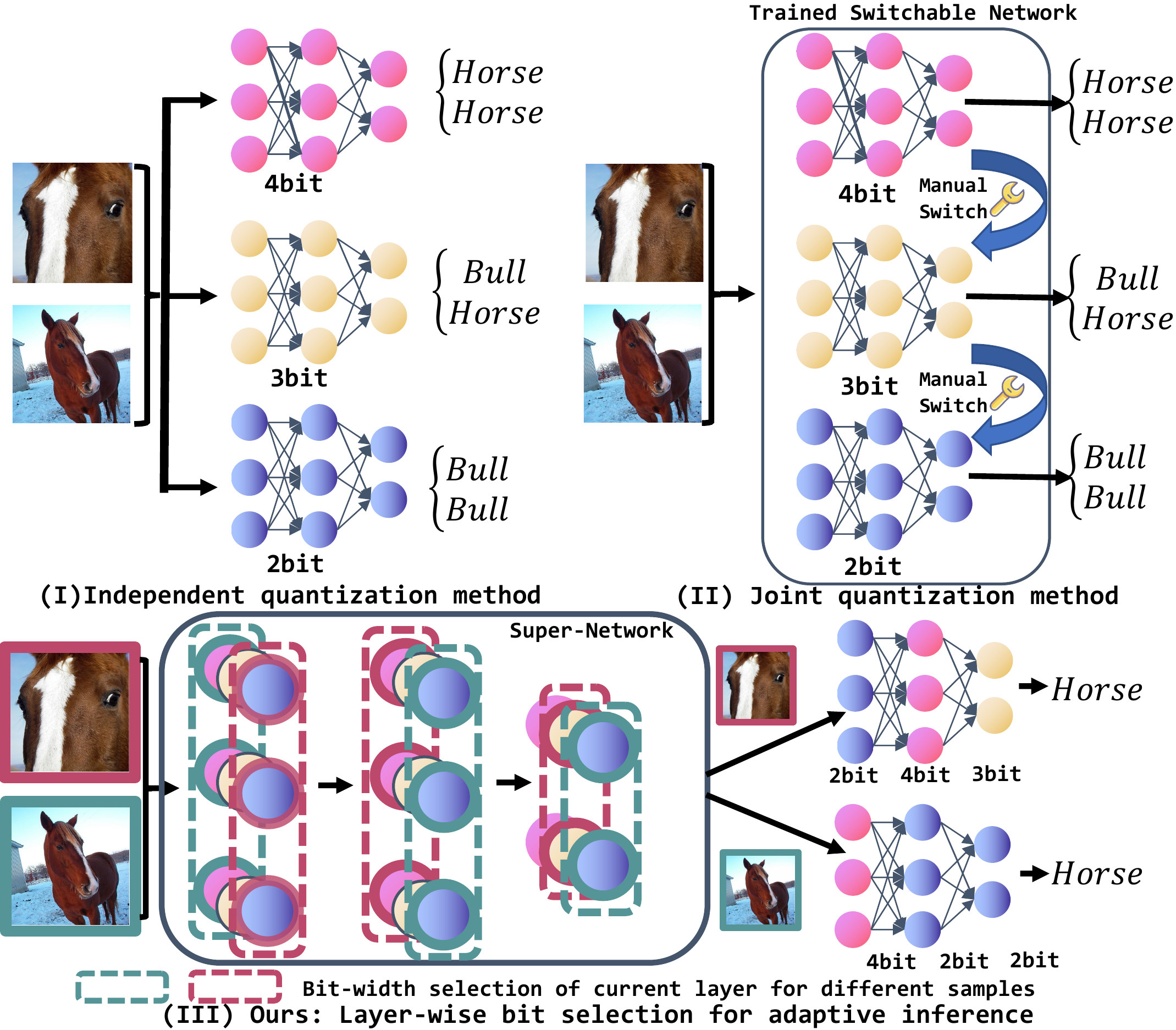}
\caption{
Inference process of a three-layer toy network with two input samples under different methods.
\textbf{(I)} 
Independent method cannot directly perform bit-width switching without retraining. 
\textbf{(II)} 
Joint method gets a switchable network that can switch bit-width of the entire network manually during inference.
They fail to exploit a fine-grained layer-wise quantization and cannot provide a data-dependent adaptive inference scheme.
\textbf{(III)} 
Our method enables adaptive inference by training a super-network that can switch bit-widths per layer and by constructing a runtime inference framework to select bit-widths to each layer according to different input data. 
It can perceive the sample differences during inference and provide a fine-grained bit-width adjustment.
}
\label{fig0}
\end{figure}

Independent methods (a.k.a quantization-aware training) \cite{zhou2016dorefa,esser2019learned,wang2019haq} tie the model to a specific quantization scheme during training, making it impossible to change the bit-width of a quantized model during inference without costly retraining. 
On the other hand, joint methods \cite{jin2020adabits,bulat2021bit,guerra2020switchable,du2020quantized,yu2019any} allow the \emph{entire} network to be switched to other bit-widths manually without retraining. 
However, they fail to fully exploit different bit-width sensitivity at the layer level, but ignore different layers that have different sensitivities to quantization \cite{dong2019hawq, wang2019haq}. 
For instance, \cite{cai2020rethinking} finds in the Inception module of the GoogleNet \cite{szegedy2015going}, the first 1$\times$1 kernel size layer has almost the same amount of computation as the last 5$\times$5 kernel size layer, but the former causes about 2\% more absolute accuracy degradation than the latter at low bit-width quantization.
This reveals that it is not necessary to use the same high bit-width for all layers of a network; instead, it is vital to use high bit-width only for those layers that are quantization sensitive, while let other layers use low bit-width as a way to achieve better efficiency.

It has been widely observed that different inputs require different computational consumption \cite{huang2018multi,wu2018blockdrop,rao2018runtime}, which is caused by the inherent ``recognition difficulty'' differences between various inputs. 
For example, an image with a clear and central object should use less computational resource (or bit-widths, in our case) than another one with a blurred object located at the edge. 
However, classical quantization methods cannot adjust the bit-width adaptively to fit this observation.
Independent methods can only use a fixed quantization scheme for all samples, because this scheme is usually assigned before training and cannot be changed during inference.
Joint methods can switch the bit-width of the entire network, but requires manual operation.
In other words, they still cannot perceive the sample differences, so they are not adaptive inference methods either.
Besides, and the most important thing is, that joint methods adjust the bit-width at a network level instead of at a praised layer level.
Modern networks generally have dozens or even hundreds of layers.
As we discussed above, the granularity of bit-width switching for the entire network is too coarse, which undoubtedly causes efficiency loss.
Therefore, to achieve better efficiency and accuracy trade-off, our \emph{core idea} is to adaptively feed different data samples with different quantization schemes at a layer level during inference. 
For example, on some layers, only samples that are difficult to recognize are assigned high bit-widths to ensure prediction accuracy.
In contrast, those samples that are easy to recognize are assigned low bit-widths to reduce computation overhead.
The difference between the classical quantization approaches and ours is illustrated in \refig{fig0}. 
The major challenges to realize this come from two perspectives. 

\textbf{(1) Exponentially increasing training space:}
Consider an $L$-layers network with $n$ optional bit-widths per layer, the number of possible quantization schemes (combinations of layers and bit-widths) is $n^L$, corresponding to $n^L$ \emph{subnets}, while the joint methods only have $n$ subnets.
For example, a ResNet34 with 4 bit-width options = $\{2,3,4,8\}$ per layer can generate $4^{34} \approx 2.9 \times 10^{20}$ potential subnets in our training space. 
Obviously, it is impractical to train so many subnets separately, as it takes several GPU days to train only a single subnet \cite{zhu2018benchmarking}, apart from the unacceptable storage overhead.
A feasible solution is to train a single weight-shared ``super-network'' that contains all subnets, rather than training all networks individually. 
However, training in such an exponential space is non-trivial, as the space is too huge to be optimized effectively. 
As we will discuss later, a simple incremental version of the previous joint methods \cite{jin2020adabits,bulat2021bit} triggers a severe accuracy degradation due to a dramatic increase in the number of subnets.

\textbf{(2) Exponentially increasing decision space:} Even with a weight-shared network, it is still challenging to determine the optimal bit-width for each layer during inference, since the decision space also grows exponentially with the deepening of layers. Simple brute force searching or random sampling to select subnets leads to sub-optimal performances, because of its excessive time complexity or its obliviousness to different input data. Naively training a decision network by collecting the accuracy and computation cost under different bit-width configurations offline is also not feasible, considering the complexity of this problem.

In this paper, we present the first work to efficiently train a layer-wise quantizable network with adaptive ultra-low bit-widths during inference. Concretely, we divide our core idea into two tractable subproblems in the \emph{training space} and \emph{runtime decision making space}, corresponding to the two challenges mentioned above.

To efficiently train the super-network that can support multiple bit-widths at the layer level, we carefully analyze the most important factors affecting the performance of the super-network, and introduce two magic codes to train it effectively. 
We further propose two key techniques called \emph{knowledge ensemble} and \emph{knowledge slowdown} to stabilize the training process, resulting in a meaningful performance improvement. 

To determine the proper configurations of each layer at runtime for various inputs, we model the optimal bit-width selection problem as a Markov Decision Process (MDP), and build a deep reinforcement learning (DRL) framework to make the online decisions under different inputs.
By this means, we are able to solve the problem that classic quantization methods cannot perceive the sample differences.

In summary, our contributions are as follows:
\begin{itemize}
\item
We propose a novel approach to train a layer-wise quantizable super-network, which only stores a \emph{single} model (i.e., the weights of different bit-widths are derived from the same stored weights, rather than stored independently) that can switch to arbitrary bit-widths at runtime for any layer. 
This greatly increases the network's runtime flexibility, providing a foundation for input-aware dynamic inference without loss of accuracy. 
Compared to joint training \cite{jin2020adabits,du2020quantized,yu2019any,guerra2020switchable}, the top-1 accuracy on ImageNet classification improves by up to 4.1\%, and we achieve that in a much more hard-to-optimize training space that is $n^{L-1}$ times larger than them.
\item
We propose a DRL-based framework to pick input-aware subnets from the trained super-network. 
The bit-width selection decision of each layer is modeled as an MDP.
Accordingly, we train a DRL agent (a very lightweight network) that can achieve adaptively inference strategy to select the bit-width of each layer to reconcile the trade-off according to different inputs. On ImageNet classification, we improve 1.1\% top-1 accuracy while using only 63.8\% BitOps compred to the data-independent quantization scheme AutoQ \cite{lou2019autoq}.
\end{itemize} 

\section{Related Work}
\subsection{Neural Network Quantization} Neural network quantization is effectively used to reduce the model storage and running overhead. Some are concerned about training a ultra-low precision model by using uniform quantization bit-width across the entire network \cite{zhou2016dorefa,baskin2021nice,esser2019learned,zhang2018lq,li2016ternary,rastegari2016xnor,choi2018pact}.
Others focus on using mixed-precision quantization for different layers. That is, the bit-widths of each layer are not exactly equal. Since different layers always exhibit different redundancy, that can greatly improve the performance of the network, avoiding forcing less sensitive layers to use higher bit-widths \cite{wu2018mixed,wang2019haq,guo2020single,dong2019hawq}. All their work already determines the bit-width of the network during training. Without retraining, it is \emph{impossible} to switch the bit-width \emph{during inference}.

\subsection{Dynamic Neural Networks} Dynamic neural networks are a type of neural networks that can change their architectures in response to different inputs. Since not all input samples require the same amount of computation to produce plausible prediction results, the early-exit mechanism is proposed in \cite{huang2018multi,fang2020flexdnn,kaya2019shallow,teerapittayanon2016branchynet}. This allows easy-to-compute samples to produce prediction results in the front layer of the network, thus avoiding additional computational consumption in subsequent layers. \cite{wu2018blockdrop, veit2018convolutional, wang2018skipnet,shen2020fractional} propose a more flexible way of dynamically adjusting the computational graph by using either a controller or a decision gate to decide block by block whether to skip it or execute it (with full-precision or lower bit), rather than skipping all layers after a decision point directly. Corresponding to the dynamic adjustment of the network depth (number of layers) is the dynamic adjustment of the network width (number of channels) \cite{chen2020storage,liu2017learning,lin2017runtime}, which is due to the fact that CNNs usually have enough redundancy in the channels to allow different pruning strategies to be generated at runtime based on different inputs. 
\subsection{Weight-Shared Networks}
Weight-shared networks \cite{jin2020adabits,yu2019universally,bulat2021bit,guerra2020switchable,du2020quantized,yu2019any} use a single set of weights to support multi-scale inference or flexible deployment without storing separate models. 
Unlike dynamic neural networks, during training, such networks are usually not constrained by the computational resources of deployment time and are therefore more flexible. The works in \cite{jin2020adabits,du2020quantized,yu2019any,guerra2020switchable,bulat2021bit} that focus on quantization are most similar to ours, in which the bit-width of their trained networks can be switched at inference time without retraining. Nevertheless, \cite{jin2020adabits,du2020quantized,yu2019any,guerra2020switchable} use the same bit-width for the whole network, ignoring that layer’s sensitivity to quantization is quite different. In other words, they do not have the ability of \emph{runtime mixed-precision}. Bit-Mixer \cite{bulat2021bit} trains a meta network with the ability of layer-wise switchable quantization level but treats all subnets equally during training, while the huge variability in convergence speed between subnets can lead to convergence to sub-optimal eventually. Consequently, each of the specific subnets requires tens of epochs for fine-tuning on the full training set to recover to normal accuracy, which is also a common drawback of weight-shared networks \cite{cai2019once,yu2020bignas}. Considering a large dataset like ImageNet, the fine-tuning time for just a single subnet can take tens of GPU hours. 

In this paper, we have carefully analyzed the most important factors affecting the weight-shared network performance and discovered a new training method. In this way, the accuracy of runtime mixed-precision on ImageNet classification can be improved significantly, the performance of our super-network can even reach the level of the separately trained networks. Moreover, the empirical results also show that the specific subnets no longer require costly fine-tuning to recover accuracy.

\begin{figure}[htbp]
\setlength{\belowcaptionskip}{-0.5cm}
\centering
\includegraphics[height=5.0cm,width=8.0cm]{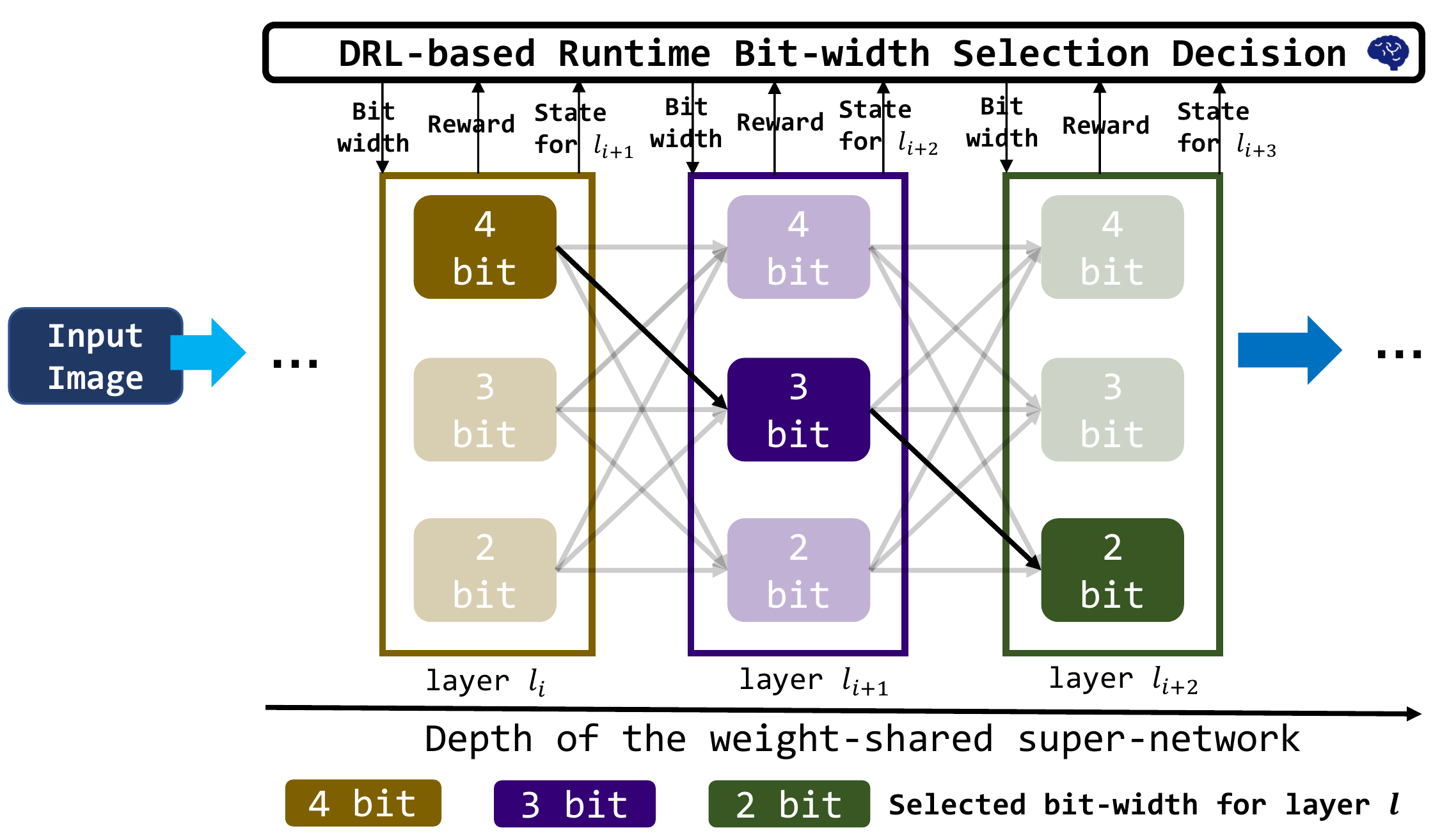}
\caption{The overall framework of ABN.}
\label{_framework}
\end{figure} 
\section{Our Approach}
The overall framework of ABN is shown in \refig{_framework}. Briefly, the weight-shared super-network is composed of a large number of subnets with the combination of layers and corresponding bit-widths. Each of the subnets can serve as a candidate for the given input data. Based on that, the DRL agent makes runtime bit-widths selection decisions layer-by-layer to determine the picked subnet for different input data, fully exploiting the advantage of ultra-low bit-widths quantization and dynamic inference.

In this section, we first discuss how to train the layer-wise quantizable super-network that supports runtime layer-wise granularity bit-widths allocation, with only one \emph{single weight-shared} model to be stored. During inference, each layer of the super-network can be allocated an ultra-low bit-width ($\leq$ 8) to construct a specific subnet. Then, we devise the DRL-based framework to select bit-width for each layer at runtime dynamically. Thus, different input data can produce a series of different bit-width selections as the input-aware subnets, achieving adaptive inference consequently.
\subsection{Quantizable Super-Network Training}
\label{sec_training}

\subsubsection{Quantization Preliminary}
\begin{figure}[t]
\begin{center}
\includegraphics[height=2.5cm,width=8.35cm]{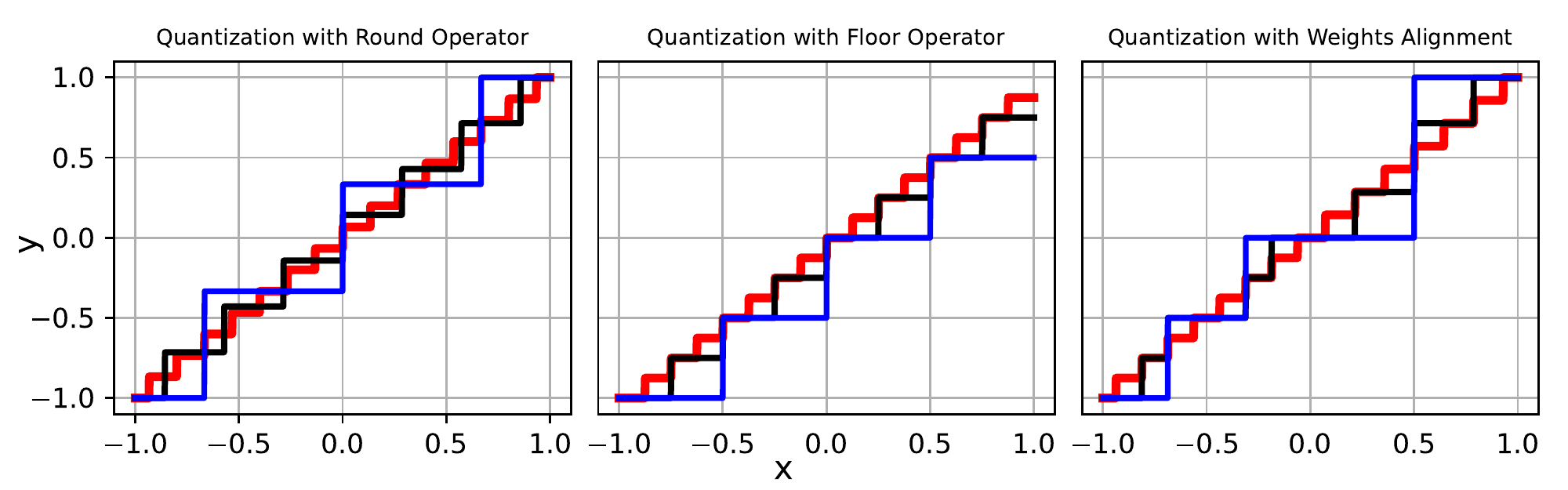}
\end{center}
\caption{Illustration of different quantization schemes.}
\label{difference_schemes}
\end{figure}

For a set of bit-widths $\B=\{b_{max}, \\b_1, ..., b_{min}\}$, we expect to find a weight-shared network that can switch each layer to any bit-widths $b \in \B$ at runtime. Namely, the weights $W$ and activations $x$ of a certain layer are both quantized to $\widehat{W_b}$ and $\widehat{x_b}$ under $b$ bit-width. To this end, we extend the traditional independent quantization training method LSQ \cite{esser2019learned}. The weights are quantized with:
\begin{subequations}
\setlength{\abovedisplayskip}{2pt}
\setlength{\belowdisplayskip}{2pt}
		\begin{align}
			&\widehat{W}_b=s_b^W \times \nint{clip(\frac{W}{s_b^W},Q_b,P_b)} \label{_eq2}\\
			&\widehat{W}_b=s_b^W \times  \nint{clip(\frac{\widehat{W}_{b_{max}}}{s_b^W},Q_b,P_b)} \label{_eq3},
		\end{align}	
\label{_eq1}
\end{subequations}
where \equref{_eq2} (Round Operator) requires the network to be stored as \emph{full-precision} but guarantees sufficient accuracy, and \equref{_eq3} (Weights Alignment) allows the network to be stored directly in bit-widths of $b_{max}$ but with a small loss of accuracy. 
For \equref{_eq3}, $\widehat{W}_{b_{max}}=W$ when $b=b_{max}$, which means the weights are obtained from the \emph{full-precision} $W$ only when $b=b_{max}$. The difference between these two formulas and the Floor Operator used in \cite{jin2020adabits,bulat2021bit} is illustrated in \refig{difference_schemes}.

Beyound that, for activations, we use:
\begin{equation}
\setlength{\abovedisplayskip}{2pt}
\setlength{\belowdisplayskip}{2pt}
\widehat{x_b}=s_b^x \times \nint{clip(\frac{x}{s_b^x},Q_b,P_b)} \label{_eq4}.
\end{equation}
Specifically, $s_b^W$ and $s_b^x$ are the learned \emph{step size scale factor} of weights and activations that need be trained for this layer. $\nint{\cdot}$ indicates that the input value is rounded to the nearest integer. $clip(r_0, r_1, r_2)$ indicates that $r_0$ will be secured between the minimum value $r_1$ and the maximum value $r_2$. Given a bit-width $b$ for this layer, $Q_b$ and $P_b$ are fixed. For weights, $Q_b=-2^{b-1}$ and $P_b=2^{b-1}-1$; for activations, $Q_b=0$ and $P_b=2^{b}-1$. 

In order to solve the problem of shifting activation distribution between different bit-widths, we use a layer-wise switchable batch normalization (BN) layer \cite{jin2020adabits,yu2019universally,ioffe2015batch}. 
To be specific, we replace the original single BN layer that follows after each convolutional layer with the bit-specified BN layers.
Namely, a layer with $n$ bit-width options has $n$ BN layers corresponding to these $n$ bit-width options.
Thus, for each convolution layer, when its allocated bit-width $b_0$ switches to $b_1$, its corresponding $s_{b_0}$, $Q_{b_0}$, $P_{b_0}$ and BN layer BN$_{b_0}$ switch to $s_{b_1}$, $Q_{b_1}$, $P_{b_1}$ and BN$_{b_1}$ accordingly.

\subsubsection{Random Sampling}
Our goal here is to find a single set of weights that will support switching the quantization level of each layer at runtime in a re-training-free fashion. 
Suppose the expected weights of the ``super-network'' is $W_S$; the aggregation of all possible configurations of bit-width is $\mathcal{C}$; and each configuration corresponds to a subnet. It is obvious that we cannot train all subnets simultaneously due to the GPU memory is finite. Thus we first investigate an intuitive approach inspired by the one-shot NAS \cite{guo2020single}, which not only trains the weights $W(j)$ of naive joint training, but also appends an \emph{additional} random sampling process, i.e., randomly sampling a bit-width configuration $c \in \mathcal{C}$ at each step. This can be expressed as:
\begin{equation}
W_S=\argmin_W\E_{c\sim{U(\mathcal{C})}}[\mathcal{L}_{train}(W(j), W(c))].
\end{equation}
In this way, it is expected that the trained network will have the ability of \emph{runtime mixed-precision} (layer-wise mutable bit-widths at runtime). The result is shown in \reftbl{_tbl1}.

\begin{table}[H]
\caption{Top1 accuracy (\%) of two training strategies of ResNet18 on ImageNet. ``Mixed'' refers to randomly selecting bit-widths to each layer during inference to test the network's performance of runtime mixed-precision.}
\setlength{\tabcolsep}{1.5mm}
\begin{tabular}{l|llll}
\hline
Method                        & 4 Bit & 3 Bit & 2 Bit & Mixed \\ \hline
Independent Training (LSQ)    & 69.6  & 68.9  & 66.3  & -     \\ 
Random Sampling               & 69.0  & 68.3  & 64.7  & 66.3  \\ \hline
\end{tabular}

\label{_tbl1}
\end{table}

Although the trained network can change the bit-width during inference without retraining, it shows a significant accuracy degradation compared to independent training.
In the most serious case (i.e., 2 Bit), it has 1.6\% top1 accuracy degradation.
That suggests that it is still something more than intuition that needs to be studied. In fact, the random sampling method is more like an incremental version of \cite{jin2020adabits} that treats all subnets equally. A similar approach is used in \cite{bulat2021bit} to make the network obtain runtime mixed-precision capability. As the results show, this intuitive approach causes the network to sub-optimal converged performance. 
Therefore, the training method of super-network should be analyzed carefully, as we will discuss in \ref{sec_training_efficiency}.

\subsubsection{Analysis of Training Efficiency}
\label{sec_training_efficiency}
\begin{figure}[t]
\centering
\includegraphics[height=2.4cm,width=6.3cm]{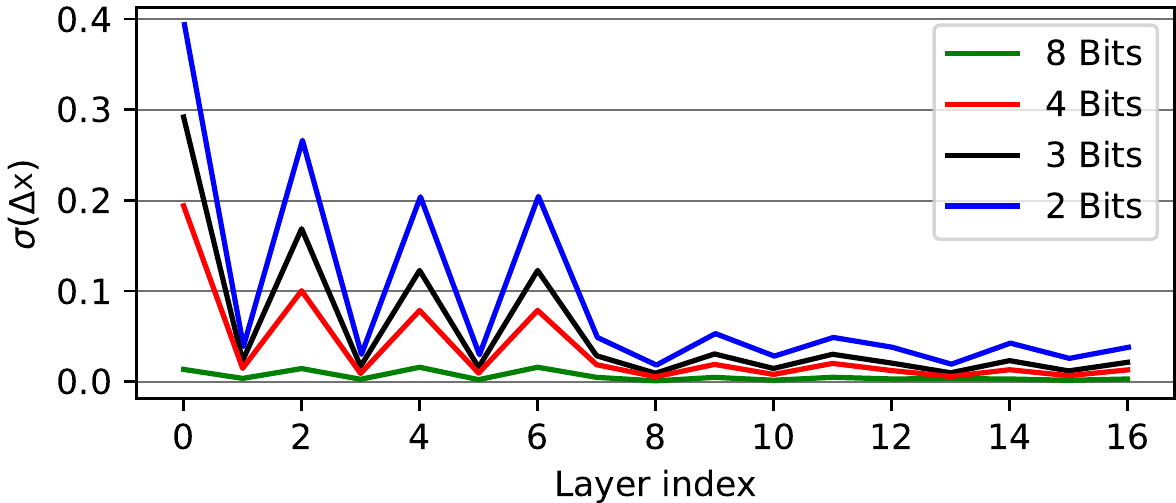}
\caption{The variance of the activations for the 16 convolutional layers of ResNet18. It can be found that the higher the bits, the smaller the variance. $\Delta x=x-\hat{x}_b$.}
\label{_fig5}
\end{figure}

Consider a convolution operation under $b$ bits:
\begin{equation}
y_b=(W+n_b^w)\odot(x+n_b^x),
\end{equation}
where $W$ and $x$ are the weights and activations, $n_b^w$ and $n_b^x$ are the quantization noise of weights and activations introduced by $b$ bits. Reducing the bit-width leads to an increase in quantization noise \cite{zhou2018adaptive}.
And as shown in \refig{_fig5}, the variance of quantized layers shows a negative correlation with bit-widths. 
Namely, as the bit-width decreases, the error increases.
Thus the absolute error of this layer can be expressed in the form of the following inequality:
\begin{equation}
|y-y_{b_{max}}|=0\leq |y-y_k|\leq |y-y_{b_{min}}|.
\label{bounder}
\end{equation}
In particular, we can deem $y\approx y_{b_{max}}$, because the highest precision output is by $y_{b_{max}}$; $y_k$ is the output of this layer under $k$ bit-width mode, $b_{min} \leq k \leq b_{max}$. The inequation indicates that performance in all subnets is bounded by the maximum and minimum bit-width mode. Optimizing the lower and upper bound can improve the accuracy of all subnets subtly. Since cross-entropy (CE) is the unmodifiable criterion of lower bound, thus the \emph{overall performance} is actually limited by the \textit{upper bound} bit-width mode $b_{min}$. That reveals the importance of the subnet whose bit-width is $b_{min}$ at each layer. That is, improving the accuracy of this \textit{crucial subnet} can potentially improve the overall performance of the entire super-network.


As shown in \reftbl{_tbl1}, \emph{runtime mixed-precision performance} of the super-network can be obtained sketchily by adding a random sampling process. Accordingly, although it is \emph{not necessary} to train all subnets at the same time, the number $k$ of random sampling is still non-trivial. Too little sampling (e.g., once) may result in some subnets not being adequately trained; too much sampling results in too much computation and may lead to intense internal conflicts within the weight-shared super-network, affecting the convergence seriously. The effect of different random sampling numbers $k$ will be further demonstrated in the experiment. 

To sum up, the \textit{two magic codes} of efficiently training a layer-wise quantizable super-network are to \textbf{improve the accuracy of the crucial subnet} and \textbf{to randomly assign bit-width per layer during training}. 
For the former, we propose two techniques called \emph{knowledge ensemble} and \emph{knowledge slowdown} to boost the accuracy of the $b_{min}$ bit-width mode. For the latter, we experimentally explore the effect of the number of random sampling on the super-network.

Notice that in order to ensure contextual consistency in this paper for clear expression, we logically divide the training process of the super-network into \textit{four continuous sub-stages} for later description, which also corresponds to different subnets, namely:

\textbf{(I)} A maximum bit-width \textit{uniform} stage \boldsymbol{$S_{max}^u$} (i.e., each layer is equally allocated the bit $b_{max}$).

\textbf{(II)} A middle bit-width \textit{uniform} stage \boldsymbol{$S_{mid}^u$} (i.e., each layer is allocated an equal bit, except for $b_{min}$ or $b_{max}$).

\textbf{(III)} $k$ random \textit{nonuniform} sampling stage \boldsymbol{$S_{rand}^n$} (i.e., the bit of each layer is randomly allocated).

\textbf{(IV)} A minimum bit-width \textit{uniform} stage \boldsymbol{$S_{min}^u$} (i.e., each layer is equally allocated the bit $b_{min}$).

\subsubsection{Knowledge Ensemble} 
\begin{figure}[t]
\centering
\includegraphics[height=2.5cm,width=6.4cm]{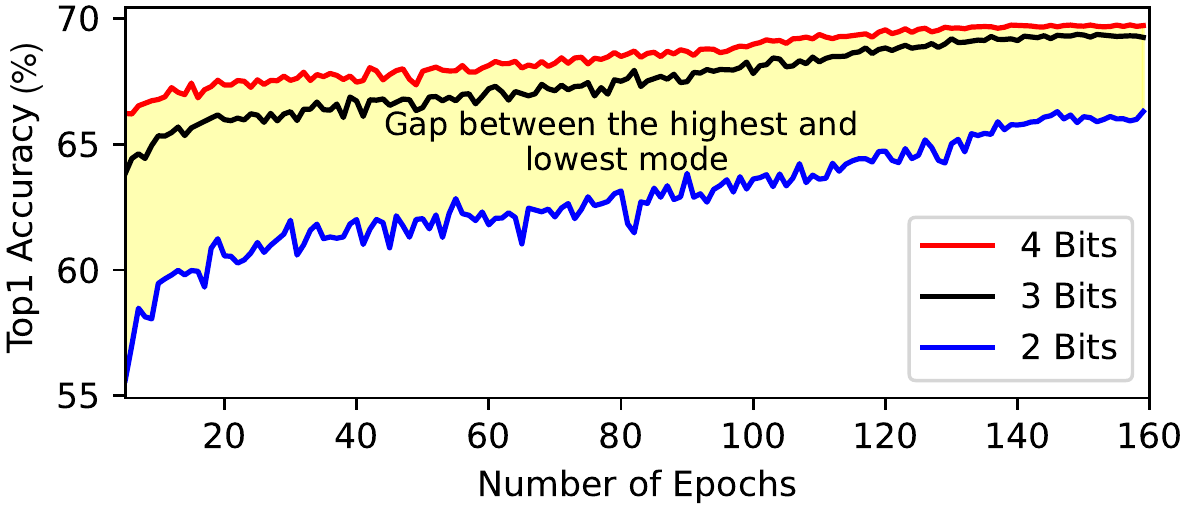}
\caption{
Asymmetric convergence rate of ResNet18 on ImageNet between different bit-widths, which shows a great gap between the 2-bits and 4-bits.
}
\label{_fig_gap}
\end{figure}

Knowledge distillation (KD) is the most famous means to train a weight-shared network \cite{hinton2015distilling,du2020quantized,anil2018large,yu2019universally}, by using soft-lables of the highest accuracy subnet (i.e., \boldsymbol{$S_{max}^u$}) as the ``teacher'' to guide other subnets (students).
It can reduce the conflict between subnets and stabilize the training. 
However, KD does not work so well for the \boldsymbol{$S_{min}^u$} due to the \textit{extremely asymmetric convergence rate} (EACR) between the maximum and minimum bit-width. 
We show that phenomenon in \refig{_fig_gap}, where the EACR remains significant (about 10\%) even after 10 epochs.
Not only that, in experiments, we even observe a much severe missdistance in the early phases, with an accuracy gap of more than 30\%.

Some researchers find that KD can lead the students to sub-optimal converged performance when the accuracy gap between teacher and students is too large \cite{mirzadeh2020improved,gao2021residual}. 
Moreover, \cite{frankle2019early, achille2017critical} have shown that the very early training time is much essential for the network, meaning such a severe gap might damage the overall performance at the essential early phases. 
Thus the naive KD is not suitable for guiding the crucial subnet anymore because \equref{bounder} shows that if its performance is damaged, the overall performance is reduced. 

It is confirmed that multiple teachers can provide rich knowledge and then generate a much well-performed student \cite{you2017learning, liu2020adaptive}. Nevertheless, they all suffered the problem of regulating the importance between soft-labels generated by different teachers with different structures. Unlike them, we have numerous subnets with the same structure, which means the different importance of soft-labels due to different structures can be totally avoided. So that we can leverage these subnets as teachers to produce ensemble knowledge for distilling \boldsymbol{$S_{min}^u$}. That is, we use a buffer $\mathcal{D}$ to store the output logits \boldsymbol{$y_{max}^u$}, \boldsymbol{$y_{mid}^u$} and \boldsymbol{$y_{rand}^n$} of \boldsymbol{$S_{max}^u$}, \boldsymbol{$S_{mid}^u$} and \boldsymbol{$S_{rand}^n$} respectively. Then, we use the average of all soft-labels in $\mathcal{D}$ to calculate the loss of \boldsymbol{$S_{min}^u$}.


\subsubsection{Knowledge Slowdown} 
To ensure knowledge ensemble could provide stable and reliable soft-labels, and further mitigate the extremely asymmetric convergence rate problem, we take inspiration from the value-based DRL algorithms \cite{van2016deep,mnih2015human}. Specifically, we introduce a \textit{target network} to produce soft-labels instead of using the one that is being trained (a.k.a the \textit{main natwork}).

The core idea of knowledge slowdown is to generate soft-labels by using a network with the same structure as the main network, but with a slower pace of parameter updates. 
In this way, the main network changes from supervising the subnets and updating weights simultaneously to only performing weights updated. 
Thus, the above two processes of supervising and updating can radically decouple, making the training more stable. 
The parameters of the target network can be updated either by exponential moving average (EMA) or by copying directly at every C-step from the main network.



Hence, the loss function for each mini-batch $\boldsymbol{(x, y)}$ is as follows:
\begin{equation}
\left\{
 \begin{aligned}
 \mathcal{L}_{max} &= \mathcal{L}_{CE}(\boldsymbol{y_{max}^u}, \boldsymbol{y}),  \\
 \mathcal{L}_{mid} &= \mathcal{L}_{KL}(\boldsymbol{y_{mid}^u},\hat{y}_{max}),  \\
 \mathcal{L}_{rand} &= \mathcal{L}_{KL}(\boldsymbol{y_{rand}^n},\hat{y}_{max}),  \\
  \mathcal{L}_{min } &= \mathcal{L}_{KL}(\boldsymbol{y_{min}^u},\frac{1}{\left|\mathcal{D}\right|}\sum\nolimits_{i\in \mathcal{D}}\boldsymbol{\hat{y_i}}),
 \end{aligned}
\right.
\end{equation}
where $\widehat{y}$ denote the soft-labels produced from the $\textit{target network}$ under different modes, $\mathcal{L}_{CE}$ is the cross-entropy (CE) loss and $\mathcal{L}_{KL}$ is the kullback–leibler (KL) loss. 



\subsection{Runtime Layer-wise Bit-width Selection}
\label{sec_decision}
After obtaining the super-network, we start to consider the problem of making the layer-wise bit-width selection decision based on different input samples at runtime. 
In general, there are two ways to achieve this: the first way is to perform a one-time decision, i.e., for an input sample, a vector $\Vec{v}$ is output at once, with each of its components corresponding to the bit-width of each layer; the second way is to carry out a step-by-step decision, i.e., the bit-widths are selected layer by layer for an input sample. 
Since the impact from quantization accumulates as the layers go deepened, we propose using the second way so that each decision is made sequentially.

For a given layer $l_i$, we want the bit-width $b$ for weights $W_i$ and activations $x_i$ that can achieve higher accuracy and lower computational consumption, which can be formed as the following objective:
\begin{equation}
\begin{aligned}
\min_{b}\E[\mathcal{L}_{T}({q(W_i, b) \odot q(x_i, b)})-\mathcal{L}_{C}(l_i, b)], \\
\end{aligned}
\end{equation}
where $q(a,b)$ is the quantization function mentioned in \equref{_eq1} and \equref{_eq4} that quantizes the input tensor $W_i$ or $x_i$ to $b$ bit-width, $\odot$ is the convolution operation, $\mathcal{L}_T$ is the loss of the task (e.g., cross entropy), and $\mathcal{L}_C$ is the computational costs in this layer at bit-width $b$ (e.g., BitOps).

For an $L$-layers neural network with $n$ bit-widths for each layer, the time complexity of gathering the supervised configurations is $\mathcal{O}(\prod_{i=1}^Ln)$. Due to this exponential time complexity, the objective function cannot be optimized directly using conventional supervised learning for the existing deep neural network with dozens or tens of layers.

To solve this, we first model the choice of optimal bit-width as a layer-by-layer MDP. 
Then, we build a DRL-based framework to make the step-by-step bit-width selection decision. The details of MDP are as follows.
\subsubsection{State}
We construct the state as an embedding vector, which consists of three parts as follows: \textbf{(I)} A fixed-length vector of input feature map of current layer. For the input feature map $\mathbf{F}_i \in \R^{c^{in}_i \times w_i \times h_i}$ of layer $l_i$, we first use the global pooling to make its dimension to $\R^{c^{in}_i}$, where $c^{in}_i$, $w_i$ and $h_i$ is the input channel number, width and height of layer $l_i$. After that, for different input channel number of different layers, we then use a fully-connected layer to project the pooled feature into a fix-length vector $\Vec{f_i}$. \textbf{(II)} Layer index $i$. \textbf{(III)} The action of last layer $a_{i-1}$.
\subsubsection{Action}
The action $a$ is defined as the bit-width for layer $i$. 
Since we are mainly concerned with ultra-low bit-width ($\leq$ 8) and employ a layer-by-layer approach, a discrete action space is enough to determine the bit-width of each layer.

\subsubsection{Reward Shaping}
The reward should consider the accuracy and computational consumption of the super-network. Therefore we define $\mathcal{R}_T$ as the final accuracy of the task, and we expect it to be as high as possible, $\mathcal{R}_C(l_i, b)$ is the computational consumption (BitOps) of layer $i$ under $b$ bit-width, where we prefer it to be as low as possible. So the reward of action $a_j$ for $i$-th layer is defined as:
\begin{equation}
r(a_j)=\left\{\begin{array}{l}
\mathcal{R}_T - \alpha \times \mathcal{R}_C(l_i,a_j), \textrm{if the last layer (i = L)},
\\-\alpha \times \mathcal{R}_C(l_i,a_j), \textrm{otherwise}
\end{array}\right.
\end{equation}  
where $\alpha$ is the hyper-parameter that drives the trade-off between accuracy and computational consumption.
To decide the action under current state $s_i$ for layer $i$, we leverage a Q-learning \cite{mnih2015human} method that define a \textit{action-value function} of expected reward under certain action as $Q(s_i,a_j;\theta)$, where $\theta$ indicates the parameters of DRL agent. Then each optimal action $a_t$ for layer $i$ is the action that maximizes the \textit{action-value function}, which can be described as $a_t=\arg\max_{a_j}Q(s_i,a_j;\theta)$. The loss function of the DRL agent can be formed by the Bellman equation:
\begin{equation}
\mathcal{L}(\theta)=(r + Q(s_{i+1},\arg\max_{a_j}Q(s_{i+1},a_j;\theta);\theta') - Q(s_i,a_t;\theta))^2.
\label{qlearning0}
\end{equation}

Thus in our DRL framework, an input image will generate a series of states corresponding to the layers to be decided.

\section{Experiments}
In this section, we first evaluate the performance of the consistent training algorithm of the super-network on ImageNet classification. Next, we conduct experiments on the DRL-based runtime bit-width selection.
We conducted experiments of ResNet18/34/50 \cite{he2016deep}, and a compact architecture MobileNet \cite{howard2017mobilenets} on ImageNet 2012 \cite{russakovsky2015imagenet} to verify the performance of ABN.

\subsection{Implementation Details}
For the super-network, we use the pre-trained model as initialization, and we keep the first and last layer at full-precision \cite{zhou2016dorefa}. 
All ResNet models are trained for 160 epochs and MobileNet is trained for 130 epochs, both using the cosine scheduler and the SDG optimizer. The initial learning rate is 0.02 for all ResNet models, 0.01 for MobileNet. The weight-decay for all models is $10^{-4}$.
We use the method in \cite{bhalgat2020lsq+} to initialize the step size factor of weights. 
We use the basic data augmentation method. All training data are randomly cropped to 224$\times$224 and randomly flipped horizontally. 
The number of random sampling $k=2$. The parameters of the target network are updated by EMA from the main network, as EMA generally ensures the stability of RL training \cite{srinivas2020curl}.

The bit-width options are \{4, 3, 2\} for ResNet and \{8, 6, 4\} for MobileNet.
These options are considered the fact that ultra-low bit-width ($\leq$4) quantization is much more difficult than high bit-width ($>$4), therefore if our method is available in the ultra-low bit-width it also can be generalized to higher bit-widths. 

We observe the same non-convergence problem as Adabits \cite{jin2020adabits} when weights directly are quantized by using \equref{_eq3} for $b_{min}$ \textless 3. 
Adabits addressed this by storing the weights to full-precision. 
To take a step further, we add an 8 bit-width mode as \boldsymbol{$S_{max}^u$} and then clip all subnets containing 8 bits after convergence, which reduces the storage footprint $4\times$ but causes a bit of degradation of accuracy. 
For a fair comparison, we provide the FP results trained by \equref{_eq2}.

The DRL agent is a very lightweight network with only 5 fully-connected layers, each with between 64 and 256 neurons.
As a comparison, FLOPs of the DRL agent and ResNet18 are 0.15M and 1819M, respectively.
To save training time, we sampled 10\% data of ImageNet2012 training set for training the DRL agent. 
It is trained by the Adam optimizer with a learning rate of $10^{-6}$. 

\begin{table*}[t]
\caption{Top-1 accuracy(\%) on ImageNet for different super-networks. We implement both \equref{_eq2} and \equref{_eq3} for training, where \equref{_eq2} ensures higher accuracy and \equref{_eq3} reduces the storage footprint with a tiny accuracy degradation. The best results overall are bolded in each metric, and the underline is the best result of all baselines. ``-'' indicates that the ``Mixed'' mode is not supported. 
Please note that \underline{``Independent Training''} is the result of separate training of different bit modes. We provide this result to prove that \underline{only our training method meets or even exceeds the performance of separate training}. 
}
\centering
    \setlength{\tabcolsep}{1.4mm}{
    \begin{tabular}{c|c|c|c|c|c|c|c|c}
    \hline
Network                        & Bit Mode   & \textbf{Ours} & \textbf{Ours (\equref{_eq3})} & Bit-Mixer \cite{bulat2021bit} & Adabits \cite{jin2020adabits} & APN \cite{yu2019any} & FQDQ \cite{du2020quantized}  & Independent Training \\ \hline
\multirow{4}{*}{ResNet18}      &  4         & \textbf{69.8} & 68.9        & \underline{69.2} \dlta{-0.6}                  & \underline{69.2} \dlta{-0.6}    & 67.9 \dlta{-1.9} & 66.9 \dlta{-2.9}           & 69.6 \\
                               &  3         & \textbf{69.0} & 68.6        & \underline{68.6} \dlta{-0.4}                  & 68.5 \dlta{-0.5}    &                  & 66.2 \dlta{-2.8}                       & 68.9\\
                               &  2         & \textbf{66.2} & 65.5        & 64.4 \dlta{-1.8}                  & \underline{65.1} \dlta{-1.1}    & 64.1 \dlta{-2.1} & 62.1 \dlta{-4.1}                       & 66.3\\
                               &  Mixed     & \textbf{67.7} & 66.5        & \underline{65.8} \dlta{-1.9}                  & -                   & -                & -                                      & -    \\  \hline
\multirow{4}{*}{ResNet34}      &  4         & \textbf{74.0} & 73.5        & 72.9 \dlta{-1.1}                  & \underline{73.5} \dlta{-0.5}    &                  &                                        & 73.8  \\
                               &  3         & \textbf{73.3} & 73.0        & 72.5 \dlta{-0.8}                  & \underline{73.0} \dlta{-0.3}    &                  &                                        & 73.0 \\
                               &  2         & \textbf{71.7} & 70.3        & 69.6 \dlta{-2.1}                  & \underline{70.4} \dlta{-1.3}    &                  &                                        & 71.1  \\
                               &  Mixed     & \textbf{72.5} & 71.6        & \underline{70.5} \dlta{-2.0}                  & -                   &                  &                                        & -    \\ \hline
\multirow{4}{*}{ResNet50}      &  4         & \textbf{76.8} & 76.2        & 75.2 \dlta{-1.6}                  & \underline{76.1} \dlta{-0.7}    & 74.9 \dlta{-1.9} &                                        & 76.6 \\
                               &  3         & \textbf{76.2} & 75.1        & 74.8 \dlta{-1.4}                  & \underline{75.8} \dlta{-0.4}    & 74.5 \dlta{-1.7} &                                        & 75.8   \\
                               &  2         & \textbf{74.3} & 73.5        & 72.1 \dlta{-2.2}                  & \underline{73.2} \dlta{-1.1}    & \underline{73.2} \dlta{-1.1} &                            & 73.5 \\
                               &  Mixed     & \textbf{75.5} & 74.3        & \underline{73.2} \dlta{-2.3}                  & -                   & -                &                                        & -   \\ \hline
\multirow{4}{*}{MobileNetV1}   & 8          & \textbf{72.6} & 72.5        &                                   & \underline{72.3} \dlta{-0.3}    &                  &                                        & 72.7   \\
                               & 6          & 72.2 & 72.2        &                                   & \textbf{72.3} \dlta{+0.1}    &                  &                                        & 72.3 \\
                               & 4          & \textbf{70.7} & 70.6        &                                   & \underline{70.4} \dlta{-0.3}    &                  &                                        & 70.7  \\
                               & Mixed      & \textbf{71.2} & 71.1        &                                   & -                   & -                &                                                    & -     \\ \hline
\end{tabular}
    }

    \label{_tb_acc}
\end{table*}

\subsection{Results for Super-Network} In \reftbl{_tb_acc}, we report our results and compare them with other state-of-the-art quantization algorithms. Compared to independent/joint methods, our super-network surpasses the accuracy of uniform bit-width mode (i.e., the bit-width of each layer is equal) in most results. Compared to the recently proposed Bit-Mixer \cite{bulat2021bit}, we also achieve a significant improvement in accuracy. In particular, on ResNet34 and ResNet50, our training method can improve mixed-precision accuracy by 2.0\% and 2.3\%. This is a good proof that when training the super-network, as we analyzed in the previous section, the crucial subnet should be treated specially, rather than treating all subnets equally. 

Overall, we note that the top1 accuracy has improved by up to 4.1\%, and even the average improvement is about 1.36\%. In addition to the accuracy, the super-network also capable of assigning bit-widths for each layer during inference time, which provides the foundation for adaptive inference. 

\subsubsection{Fine-Tuning Subnets}
\begin{figure}[h]
\centering
\includegraphics[height=4cm,width=8cm]{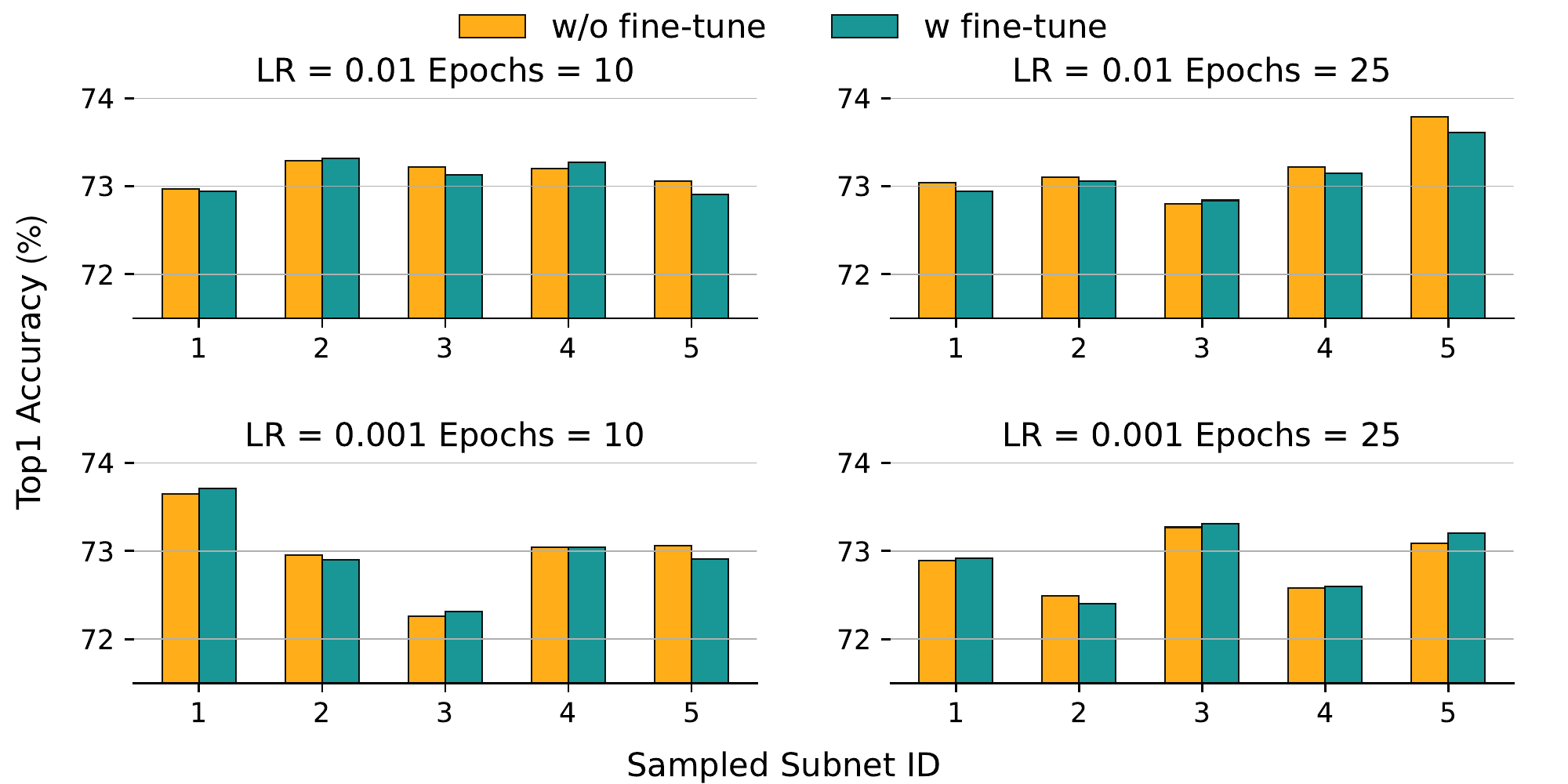}
\caption{Top1 accuracy (\%) of fine-tuning for ResNet34.}
\label{_fig_finetuning}
\end{figure}

To further verify that our training solution achieve good enough performance on specific subnets, we randomly sampled 20 subnets out of the super-network, using the fine-tuning settings in \cite{yu2020bignas} (LR = 0.01/0.001, 10/25 epochs on the full training set, etc.). 
As shown in \refig{_fig_finetuning}, the accuracy transformation of these subnets after fine-tuning fluctuates only around $\pm$0.1\%. 
This indicates that our training strategy results in relatively optimal performance for any subnets.

\subsubsection{Results for Learned Factors} 
\begin{figure}[h]
\setlength{\belowcaptionskip}{-0.4cm}
\centering
\includegraphics[height=2.5cm,width=6.5cm]{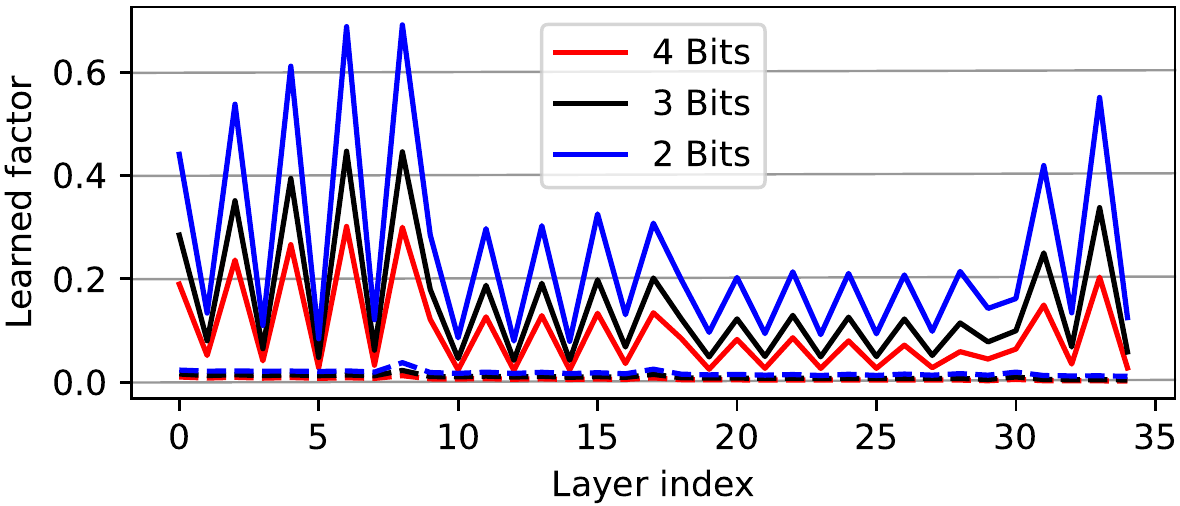}
\caption{Learned factors for weights (dashed line) and activations (solid line) of ResNet34 on ImageNet.}
\label{_fig4}
\end{figure}

\refig{_fig4} shows the learned step size scale factors (learned factors) of different layer activations and weights of ResNet34. We find that the difference in learned factors between different bit-widths is relatively large for the same layer. This illustrates the importance of using a unique factor for each bit-width. Also, for a smaller bit-width (e.g., 2bit), our training algorithm gives a larger learned factor compared to a larger bit-width (e.g., 4bit) to make the quantized values more suitable for the distribution of the smaller bit-width.

\subsubsection{Empirical Results for Random Sampling}
\begin{figure}[h]
\centering
\includegraphics[height=2.3cm,width=7.3cm]{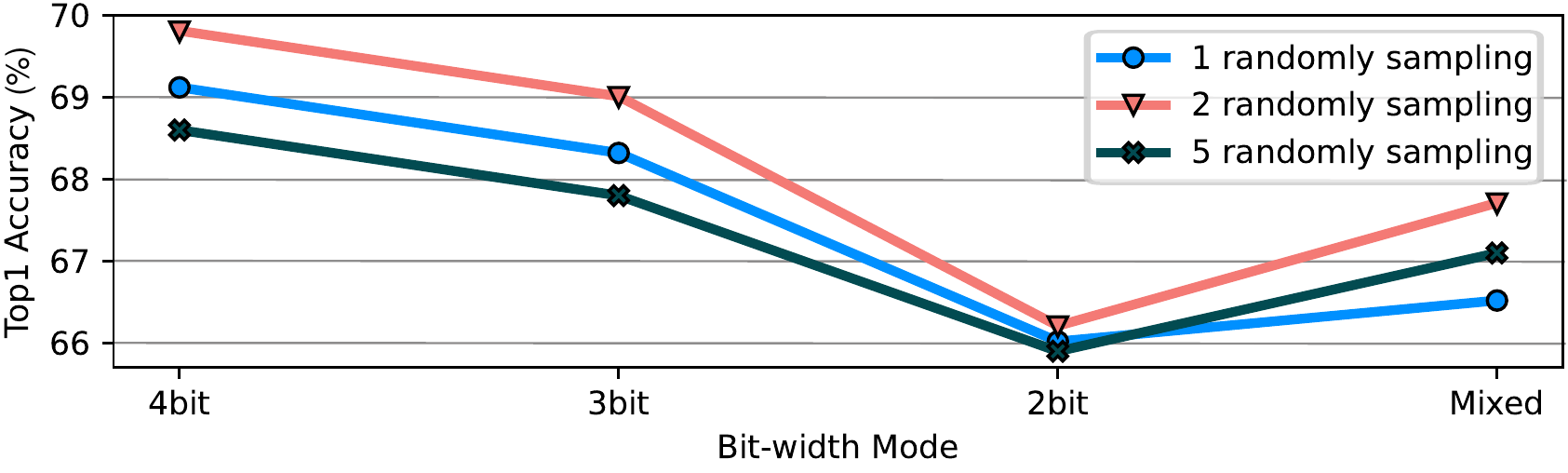}
\caption{Top1 accuracy (\%) with different $k$ of ResNet18 on ImageNet.}
\label{_fig_inrs}
\end{figure}

\refig{_fig_inrs} shows the results of random sampling $k$ with different values. 
We find that only random sampling once could lead to the runtime mixed-precision performance poorly. 
However, it is still very unwise to increase the number of random sampling heavily.
Because that augments the training overhead and causes violent conflicts between sampled subnets, resulting in the network cannot be correctly converged. 
Results show that the mixed-precision performance of super-network is not positively correlated with the sampling number.
Empirically, it has an optimal sampling number $k=2$, where the accuracy and training costs are both considered.
\subsection{Results for Runtime Layer-wise Bit-width Selection}
We verify the feasibility of the adaptive inference strategy on the ResNet18 super-network.
The results of two DRL policies in \reftbl{_drl_result} show that we achieved 0.8\% accuracy improvement while using only 77.2\% of the computational resources (BitOps) compared to baseline. Compared to the classical fixed mixed-precision scheme AutoQ, we also can achieve 1.1\% accuracy improvement with 36.2\% BitOps saving. In addition, since the DRL agent is actually a shallow network, its computation only accounts for about 2\% of the overall network overhead. By replacing the four fully-connected layers we used with RNN, the computational cost of the DRL agent can be further reduced \cite{wang2018skipnet}.

\begin{table}[t]
\caption{Result for runtime layer-wise bit-width selection. \texttt{CCR} indicates \textbf{C}onstrained \textbf{C}omputational \textbf{R}esources, which means the scenarios with limited computational resources, i.e., prioritizing fewer BitOps with as little impact on accuracy as possible. \texttt{HAR} indicates \textbf{H}igher \textbf{A}ccuracy \textbf{R}equirement, which is suitable for scenarios requiring higher accuracy. 
``Acc.'' means top1 accuracy (\%). 
``$\Delta$Acc.'' means the top1 accuracy improvement compared to the baseline.
``W'' denotes the bit-width of weights. 
``A'' denotes the bit-width of activations.
$^{\dagger}$: Layer-wise.
$^{\uparrow}$: Nice cannot switch bit-width during inference.
$^{*}$: Joint methods do not support adaptive inference, they need switch the bit-width of the entire network manually.
}
\small
\setlength{\tabcolsep}{1.1mm}
\begin{tabular}{l|llll}
\hline
Method                                        & Bit-width        & Acc.                & \Delta Acc.   & BitOps               \\ \hline
FQDQ$^*$ (baseline) \cite{du2020quantized}                   & 3W3A             & 66.2                &  0            & 100.0\%              \\
AdaBits$^*$ \cite{jin2020adabits}                 & 3W3A             & 68.5                &  +2.3         & 100.0\%               \\
Nice$\uparrow$ \cite{baskin2021nice}                    & 3W3A             & 67.7                &  +1.5         & 100.0\%               \\
APN$^*$     \cite{yu2019any}                      & 4W4A             & 67.9                &  +1.7         & 177.7\%              \\
AutoQ$^{\dagger}$   \cite{lou2019autoq}       & Static Mixed            & 67.5                &  +1.3         & $\sim$ 136.7\%       \\ \hline
Ours (DRL \& \texttt{CCR})                    & Runtime Mixed    & 67.0       &  +0.8         & \textbf{77.2}\%      \\
Ours (DRL \& \texttt{HAR})                    & Runtime Mixed    & \textbf{68.6}       &  \textbf{+2.4}         & \textbf{87.3}\%      \\ \hline
\end{tabular}

\label{_drl_result}
\end{table}

\begin{figure}[h]
\centering
\includegraphics[height=2.25cm,width=1\columnwidth]{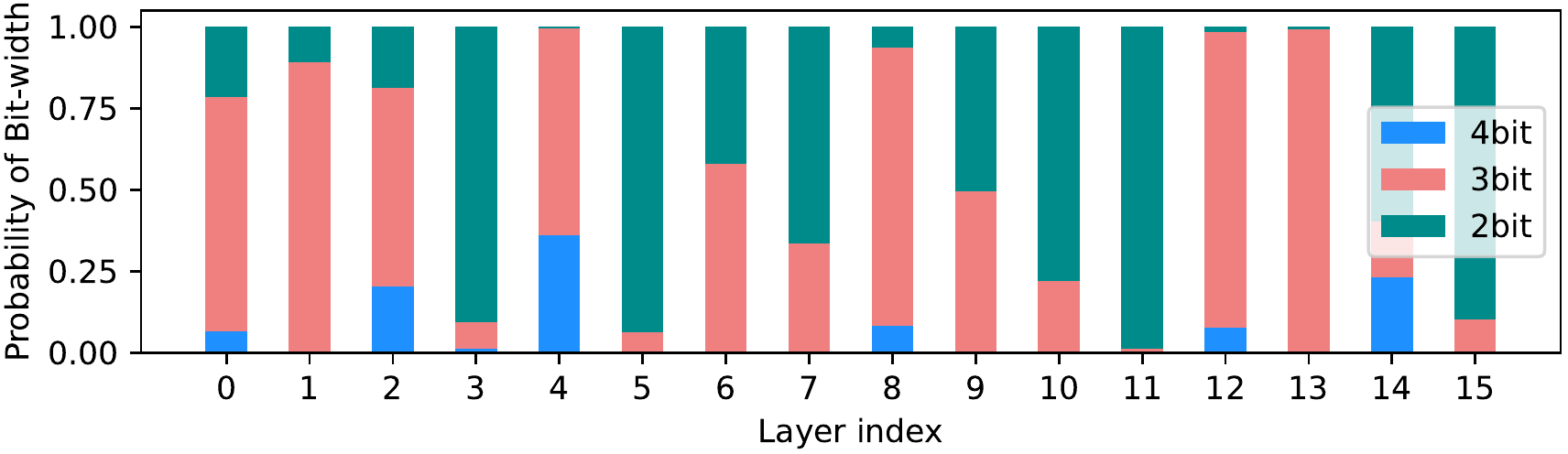}
\caption{Action (Bit-width) selection probability of ResNet18 on ImageNet for \texttt{CCR}. }
\label{_fig_prob}
\end{figure}

We show the behavior of the CCR policy in \refig{_fig_prob}. In CCR, the computational constraints are more stringent, so the DRL agent tends to allocate more low bit-widths. We observe that the DRL agent tends to use a large number of 2 and 3 bits to save BitOps. Also, the top layers have a high probability of using 4 bits, because these layers need high precision to extract low-level features. This suggests that the DRL agent adaptively takes different actions to reconcile computational consumption and accuracy for different inputs.

\begin{figure}[h]
\centering
\includegraphics[width=1.0\columnwidth]{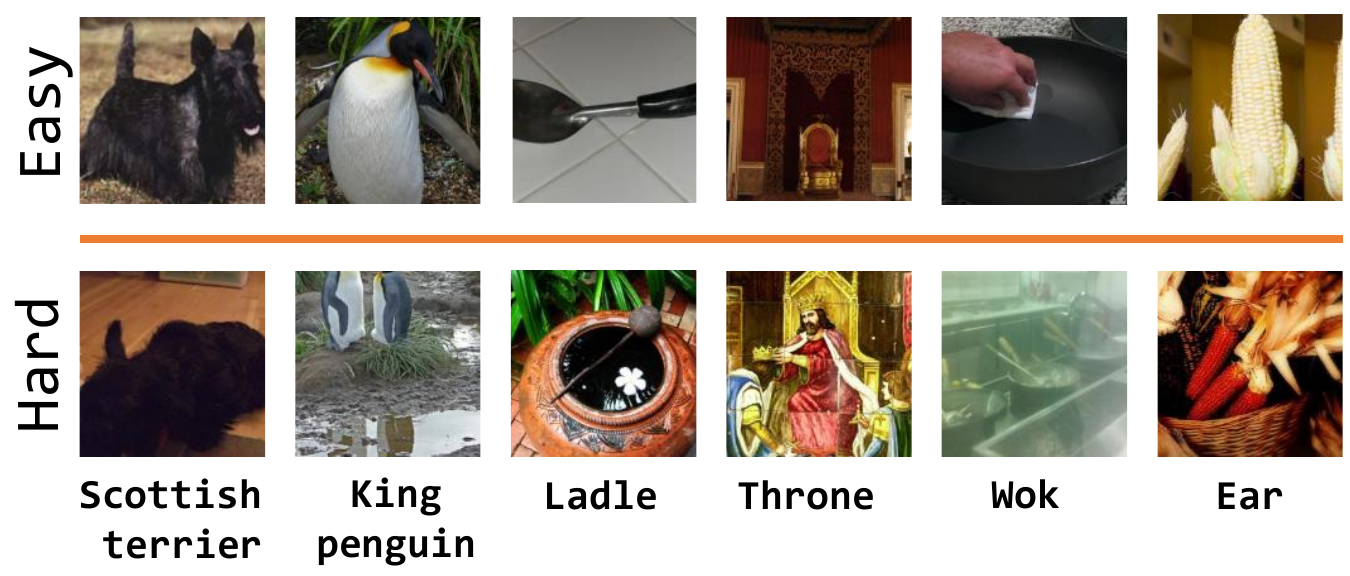}
\caption{Visualization of easy (lower bit-widths are allocated by our DRL agent) and hard (higher bit-widths are allocated by our DRL agent) samples.}
\label{_fig_vis}
\end{figure}

To further understand the behavior taken by the HAR policy, we divide the samples into two categories and visualize them in \refig{_fig_vis}, namely \textit{easy} samples (less bit-widths are allocated to save computation; about 90\% of average BitOps) and \textit{hard} samples (larger bit-widths are allocated to ensure accuracy; about 126\% of average BitOps). We find that lower bit-widths are used for clear samples or samples where the entire object appears in the image, while higher bit-widths are used for blurred samples or where the target object is at the edge of the image. 


\subsection{Ablation Studies for Knowledge Ensemble and Knowledge Slowdown}
\reftbl{_ablations} shows that knowledge ensemble can boost the accuracy of 2-Bit and mixed 1.1\% and 0.5\%, respectively. With the combination of knowledge ensemble and knowledge slowdown, the accuracy of 2-Bit and mixed can be further improved by 2.4\% and 1.3\%, respectively. 
This demonstrates the effectiveness of these two techniques, which can alleviate the training difficulties caused by the exponential growth of training space and significantly boost the performance of the super-network.

\begin{table}[h]
\caption{Top1 accuracy (\%) results of ResNet34 on ImageNet for ablation study.}
\centering
\setlength{\tabcolsep}{0.9mm}
\begin{tabular}{c|c|c|c}
\hline
Knowledge ensemble & Knowledge slowdown      &    2 Bit      &    Mixed\\ \hline
   \ding{55}       &      \ding{55}          &    69.3       &    71.2   \\
   \ding{51}       &      \ding{55}          &    70.4       &    71.7   \\
   \ding{51}       &      \ding{51}          &    71.7       &    72.5   \\ \hline
\end{tabular}

\label{_ablations}
\end{table}

\section{Conclusion}
This paper proposes the ABN to achieve layer-wise ultra-low bit-width adjustment adaptively according to specific input data. 
To do this, we solve two challenges.
The first one is how to efficiently train one network that contains multiple possible bit-widths for each layer.
The second one is how to determine the appropriate bit-width of each layer for different samples.
For the former, we find the crucial subnet that has the greatest impact on the overall performance of the super-network, and propose two key technologies to push the performance of this lower bound.
For the latter, we model the optimal bit-width selection problem as an MDP, and then propose a DRL-based adaptive inference strategy to pick input-aware subnets from the super-network.
ABN can capture the differences across various inputs and then adjust bit-width on the fly, which makes it possible to guarantee sufficient accuracy while effectively reducing computational consumption.

\bibliographystyle{ACM-Reference-Format}
\bibliography{sample-base}



\end{document}